\documentclass[10pt, conference]{IEEEtran}

\usepackage{amsmath,amssymb,amsfonts}
\usepackage{graphicx}
\usepackage{pifont}
\usepackage{float}

\usepackage{amsthm}
\usepackage{xcolor}
\usepackage[figuresright]{rotating}
\usepackage[compatibility=false]{caption}

\usepackage{graphicx}
\graphicspath{{/}{fig/}}

\usepackage{array}
\usepackage{textcomp}
\usepackage{xcolor}
\usepackage{multirow}
\usepackage{booktabs}
\usepackage{enumerate}

\usepackage{mathtools}
\usepackage{breqn}
\usepackage{float}

\usepackage{pgfplots}
\pgfplotsset{compat=1.7}
\usepgfplotslibrary{groupplots}

\usepackage{caption}
\usepackage{subcaption}

\usepackage{multirow,tabularx}
\usepackage{hyperref}
\usepackage{flushend}
\usepackage{algorithmic}
\usepackage[vlined, ruled, shortend]{algorithm2e}

\usepackage{pgfplots}
\pgfplotsset{compat=1.7}
\usepgfplotslibrary{groupplots}

\usepackage[font=small]{caption}
\usepackage{subcaption}

\usepackage{multirow,tabularx}
\usepackage{hyperref}
\usepackage{flushend}
\usepackage{algorithmic}
\usepackage{soul}

\usepackage[vlined, ruled, shortend]{algorithm2e}

\usepackage{pgfmath}
\usetikzlibrary{decorations.text, arrows.meta,calc,shadows.blur,shadings, spy}

\usepackage{listings}
\newlength\figureheight
\newlength\figurewidth
\SetAlCapNameFnt{\footnotesize}
\SetAlCapFnt{\footnotesize}

\title{
    Decentralized Intent-Based Multi-Robot Task Planner with LLM Oracles on Hyperledger Fabric
}

\author{
    \IEEEauthorblockN{
        Farhad Keramat$^{*}$\IEEEauthorrefmark{2},
        Salma Salimi\IEEEauthorrefmark{2},
        Tomi Westerlund\IEEEauthorrefmark{2}
    }
    \IEEEauthorblockA{
        \normalsize
        \IEEEauthorrefmark{2}\href{https://tiers.utu.fi}{Turku Intelligent Embedded and Robotic Systems (TIERS) Lab, University of Turku, Finland}.\\
        Emails: \textsuperscript{1}\{fakera, salmas, tovewe\}@utu.fi\\[+6pt]
    }
}

\begin{document}

\maketitle
\thispagestyle{empty}
\pagestyle{empty}

\begin{abstract}
\label{sec:abstract}

Large language models (LLMs) have opened new opportunities for transforming natural language user intents into executable actions.
This capability enables embodied AI agents to perform complex tasks, without involvement of an expert, making human-robot interaction (HRI) more convenient.
However these developments raise significant security and privacy challenges such as self-preferencing, where a single LLM service provider dominates the market and uses this power to promote their own preferences.
LLM oracles have been recently proposed as a mechanism to decentralize LLMs by executing multiple LLMs from different vendors and aggregating their outputs to obtain a more reliable and trustworthy final result.
However, the accuracy of these approaches highly depends on the aggregation method. The current aggregation methods mostly use semantic similarity between various LLM outputs, not suitable for robotic task planning, where the temporal order of tasks is important.
To fill the gap, we propose an LLM oracle with a new aggregation method for robotic task planning.
In addition, we propose a decentralized multi-robot infrastructure based on Hyperledger Fabric that can host the proposed oracle.
The proposed infrastructure enables users to express their natural language intent to the system, which then can be decomposed into subtasks. These subtasks require coordinating different robots from different vendors, while enforcing fine-grained access control management on the data.
To evaluate our methodology, we created the SkillChain-RTD benchmark made it publicly available~\footnote{https://github.com/farhadcuber/task-decomp}. Our experimental results demonstrate the feasibility of the proposed architecture, and the proposed aggregation method outperforms other aggregation methods currently in use.

\end{abstract}

\begin{IEEEkeywords}
LLM; Task Planning; LLM Oracle; Hyperledger Fabric; Blockchain;
\end{IEEEkeywords}
\IEEEpeerreviewmaketitle

\section{Introduction}
\label{sec:intro}

The recent advances in large language models (LLMs) have unlocked many opportunities in human-robot interaction (HRI). By training on vast, internet-scale datasets, LLMs can perform zero-shot reasoning on tasks they were not specifically trained for~\cite{brown2020language}. These models can perform complex logical planning and semantic decomposition without any further fine-tuning, even though fine-tuning on the specific domain can tremendously improve the accuracy of their generated results~\cite{pallagani2022plansformer}. Consequently, researchers do not view LLMs merely as linguistic tools for interpreting user input, but rather leverage them in versatile ways to open new possibilities through their broad knowledge and reasoning capabilities~\cite{bommasani2022opportunities}.

\begin{figure}[ht]
  \centering
  \includegraphics[width=0.8\columnwidth]{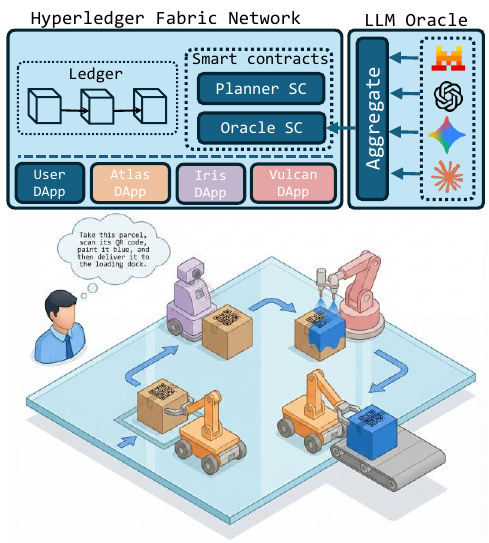}
  \caption{High-level conceptual framework of the decentralized task planning system. A natural language user intent is decomposed into robot-specific atomic tasks through a Hyperledger Fabric-based consensus across an ensemble of heterogeneous LLM oracles, ensuring a trusted and verifiable plan execution.}
  \label{fig:intro}
\end{figure}

This potential is especially transformative in the realm of robotics. Traditionally, a robot-specific expert was needed to operate the robot, one familiar with configurations unique to that robot, making a steep learning curve for adopting robots~\cite{goodrich2008human}. However, roboticists are increasingly integrating LLMs into robotic control systems, ranging from Vision-Language Models (VLM) for environmental perception to Vision-Language-Action (VLA) models for comprehensive end-to-end robot control~\cite{salimpour2025towards}. These integrations present a valuable opportunity to connect non-expert users to interact seamlessly with physical embodied agentic AIs, enabling robots to interpret complex situational contexts with a level of sophistication that was previously unattainable. These approaches range from lower-level integrations, such as using LLMs merely as an interface to existing robotic tools like ROS\,2, to high-level end-to-end multi-robot autonomy similar to the scenario depicted in Figure.~\ref{fig:intro}.

While some approaches aim to use foundation models to directly output robotic actions like VLAs, the other direction is to predefine a set of fine-grained subtasks (tools) with input and output formats and then break down the user input into these tools. These tools can be executed simultaneously or one at a time. LangChain~\cite{Chase_LangChain_2022} is one of the most well-known frameworks that provides this capability, and for example, ROSA~\cite{royce2025enabling} uses LangChain and connects it to ROS tools to enable end-to-end robotic control. In this case, it is a single agent, but similar research is also done on multi-agent and for multi-robot environments~\cite{kannan2024smart}.

While the integration of LLMs offers profound advantages for HRI, it simultaneously introduces a new surface of critical vulnerabilities that must be addressed to ensure system trustworthiness. These risks span a broad spectrum, including user data privacy concerns, threats to the physical safety of robotic platforms, and adversarial prompt injection attacks, where malicious inputs manipulate the model into bypassing safety constraints or executing unauthorized actions \cite{liu2023prompt}. 
While these security concerns are well researched and addressed, the centralization of LLM providers has not been widely studied. Relying on a single model subjects the entire ecosystem to the operational or malicious self-preferencing of a dominant provider.

The dominance of a single LLM provider, such as OpenAI, may extend beyond mere monopolization; as these models can be employed for robotic task decomposition and could be connected to real robots, they could potentially leverage physical entities in accordance with their preferences. This situation poses actual physical threat to humanity, greater risk compared to the dominance previously held by entities such as Google over search engines. An illustrative example is the European Commission's 2017 ruling against Google, which concluded that the company abused its search engine dominance to suppress rival comparison shopping services in favor of its own~\cite{GoogleAntitrust2017}.

Distributed Ledger Technologies (DLTs) offer a framework for realizing decentralized infrastructures. While the intersection of LLMs and DLTs has garnered significant research interest, few studies have explored DLTs as a primary platform for enabling truly decentralized LLM utility. One of the main challenges in using LLMs as a source of input in smart contracts is their probabilistic nature. For consensus on data, we must ensure that different nodes executing the smart contract agree on the same result. For this reason, it is not straightforward to simply call an LLM API and use the results directly.

To accomplish this consensus, new consensus mechanisms have been suggested~\cite{chen2024blockagents}, but using these new mechanisms means that we cannot easily leverage existing infrastructure such as Hyperledger Fabric, and instead must develop new infrastructure for this purpose. Another direction to tackle the probabilistic nature of LLMs is to use oracles. In this approach, LLM oracles are external entities that can validate external data and submit it to the DLT~\cite{xu2023smartllm}.

In this way, we can readily deploy, for example, a Hyperledger Fabric or Ethereum network and obtain outputs from a LLM oracle, while maintaining a modular system design. LLM oracles execute multiple instances of LLM models and subsequently aggregate their outputs. Aggregation is typically performed by comparing results pairwise and selecting the output that is most similar to the others. This similarity assessment depends strongly on the output format. Existing studies have primarily focused on general-purpose applications of LLMs, such as open-domain question answering~\cite{zeng2025connecting}, and thus the similarity measures they employ are generally based on semantic similarity or on counting the frequency of overlapping words. Since these methods have not been developed for robotic task decomposition scenarios, the resulting similarity metrics are not sensitive to the ordering of tasks. This limitation constitutes the gap that the present work aims to address. Specifically, we seek to adapt LLM oracles for robotic task decomposition and to demonstrate how they can be integrated with Hyperledger Fabric.

In this paper, we specifically investigate the utility of LLM oracles as decentralized robotic task decomposition planners, bridging the gap between non-expert natural language directives and executable robotic sub-tasks. Our primary contributions are as follows:
\begin{itemize}
    \item We propose a novel decentralized architecture that leverages an ensemble of LLM oracles to securely plan robotic tasks within a Hyperledger Fabric network.
    \item We introduce a sequence-aware aggregation method based on the Longest Common Subsequence (LCS) metric and a historical reputation mechanism to protect the integrity of decomposed task orders.
    \item We present \textbf{SkillChain-RTD}, a publicly available benchmark for robotic task decomposition, and demonstrate that our approach outperforms traditional semantic and lexical similarity methods in detecting and mitigating adversarial model behaviors.
\end{itemize}

The remainder of this paper is organized as follows: Section II provides a review of the background and related work; Section III details the proposed system architecture and methodology; Section IV presents the experimental results and the evaluation of the SkillChain-RTD benchmark; and Section V concludes the paper with a discussion on future research directions.

\section{Background and Related Work}
\label{sec:related}

\subsection{Large Language Models and Robotic Grounding} Large language models (LLMs) emerged from the development of the Transformer architecture~\cite{brown2020language}, with models such as ChatGPT being trained on internet-scale datasets to achieve general-purpose reasoning. To adopt LLMs in robotics, researchers have focused on grounding these models in physical environments. SayCan~\cite{ahn2022can} demonstrates how LLMs can be combined with pretrained low-level skills and reinforcement-learned value functions to ensure generated instructions are feasible for a robot's specific capabilities. Similarly, AutoRT~\cite{ahn2024autort} utilizes foundation models to scale autonomous data collection, though it remains limited by its reliance on scripted policies and centralized reasoning.

\subsection{LLMs as Robotic Planners} Recently, LLMs have been specifically utilized as planners for multi-agent systems. DART-LLM~\cite{wang2024dart} enables coordinated multi-robot execution by combining LLMs with Directed Acyclic Graph (DAG) dependency modeling to parse instructions and assign tasks. For long-horizon tasks, LaMMA-P~\cite{zhang2025lamma} integrates LLMs with PDDL-based search to improve success rates in heterogeneous fleets, while LLM-as-BT-Planner~\cite{ao2025llm} leverages behavior trees to enhance the robustness of assembly tasks. Comprehensive surveys~\cite{wei2025plangenllms, cao2025large} categorize these planners based on criteria such as completeness and optimality, identifying a critical need for more consistent evaluation frameworks across robotic domains.

\subsection{Benchmarks for Task Decomposition} Several benchmarks have been created to evaluate these planning capabilities. ALFRED~\cite{shridhar2020alfred} maps natural language to action sequences in household environments, while PARTNR~\cite{chang2024partnr} focuses on reasoning in human-robot collaboration. However, many existing benchmarks lack the specific granularity required to evaluate the logical decomposition of tasks for diverse industrial robot fleets. To address this, we introduce \textbf{SkillChain-RTD}, which is uniquely designed for robotic task decomposition. Unlike ALFRED or PARTNR, our benchmark focuses on the high-level mapping of intent to discrete, multi-organization skill sequences, providing a more focused dataset for testing planning logic.

\subsection{Blockchain and Oracle Integration} Blockchain serves as a foundational technology for decentralization and access control management in robotics~\cite{salimi2025abac}. The integration of LLMs and blockchain is an emerging field; BlockAgents~\cite{chen2024blockagents} utilizes a "proof-of-thought" consensus to prevent Byzantine behaviors in multi-agent systems. Research into blockchain oracles~\cite{pasdar2023connect} has classified implementations into voting and reputation-based types to bridge off-chain data with smart contracts. Frameworks like C-LLM~\cite{zeng2025connecting} and SmartLLM~\cite{xu2023smartllm} have explored using LLM oracles for semantic truth discovery and intelligent analysis within decentralized applications.

Despite these advancements, to the best of our knowledge, none of these works have utilized LLM oracles for the specific purpose of \textbf{robotic task planning}. While prior work focuses on general semantic relevance for textual data, our approach utilizes a sequence-aware aggregation method that is significantly more robust for the temporal requirements of robotic task planning.

\section{Methodology}
This section defines the problem statement and outlines the proposed methodology, which integrates blockchain technology with LLM oracles to address the identified challenges. Additionally, it examines how the proposed architecture mitigates security risks.

\begin{figure*}[t]
  \centering
  \includegraphics[width=\textwidth]{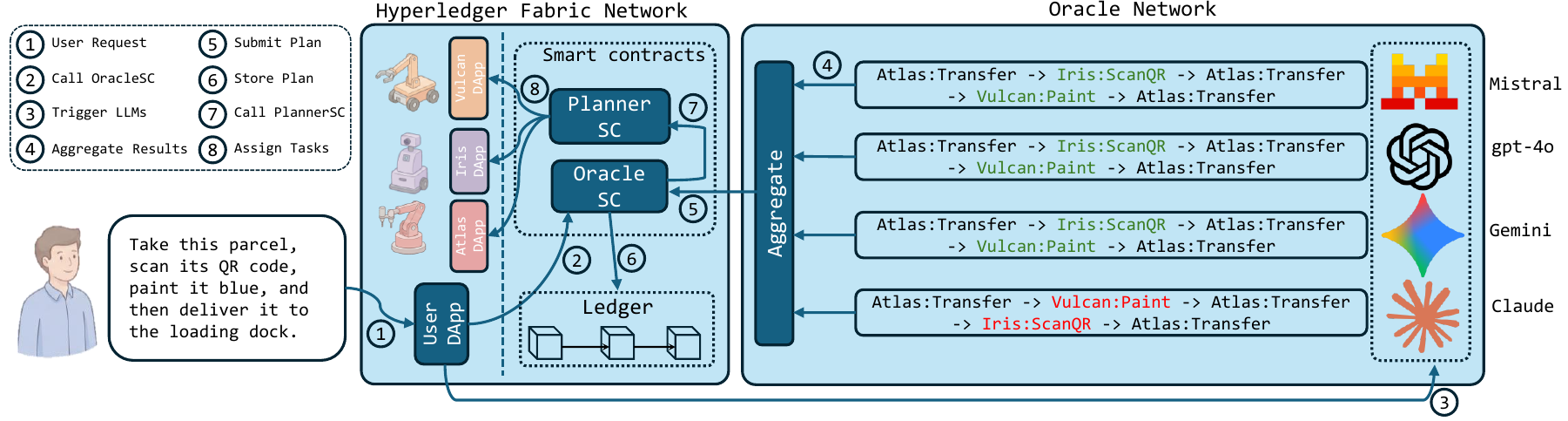}
  \caption{System architecture.}
  \label{fig:system-arch}
\end{figure*}

\subsection{Problem statement}
Consider a system with $N$ robots, denoted by $R_i$, operating under $M$ distinct organizations, $O_j$. Each robot $R_i$ is owned and serviced by exactly one organization. In addition, every robot has its own unique skill set $\mathcal{S}_{R_i}$. For simplicity, we assume that each skill is associated with only a single robot; thus, no skill is shared across the skill sets of different robots.
Users rely on these robots to fulfill specific requests, which we refer to as \textit{intents}. These intents are expressed by users in natural language and must then be translated into an executable plan. A plan is represented as an ordered list of pairs $[(R_i, s_{i}), \dots, (R_k, s_{k})]$, indicating that the corresponding tasks are to be executed sequentially. Alternatively, it is possible to model the tasks as Directed Acyclic Graphs (DAGs)~\cite{wang2024dart}, which, for the sake of simplicity, we are not considering. This planning process must be executed in a decentralized manner to ensure the correct distribution of tasks among robots and their respective organizations, which may not fully trust one another.

\subsection{System design}
LLM oracles have enabled the decentralization of LLMs by employing architectures proposed in~\cite{xu2023smartllm, zeng2025connecting}. A similar architectural design to that in~\cite{zeng2025connecting}, previously used with the Ethereum network, is adopted here with Hyperledger Fabric. The overall system design is depicted in Figure~\ref{fig:system-arch}. The Hyperledger Fabric blockchain platform is selected due to its widespread adoption in robotics~\cite{salimi2023hyperledger} and its comprehensive infrastructure, which includes smart contract execution~\cite{fu2023event} and fine-grained access control management~\cite{salimi2025abac}. In this architecture, LLM oracles operate within a separate oracle network, ensuring that only the final results are stored on the ledger. The primary contribution of this work is the aggregation module within the oracle network, which manages the merging of robotic plans as represented in the diagram.
The comprehensive workflow, from the initial user request to task execution, is structured as follows:

\newcommand{\circled}[1]{\ding{\numexpr171 + #1\relax}}
\begin{description}[\normalfont]
    \item[\circled{1}] \textbf{User Request:} The workflow begins when an authorized end user sends a natural language request through a Decentralized Application (DApp) running on the Hyperledger Fabric network.
    
    \item[\circled{2}] \textbf{Call OracleSC:} The DApp forwards this intent to the Oracle Smart Contract, which records the request on the ledger. This creates a transparent audit trail and enables the contract to asynchronously link upcoming LLM responses to the initial user query.
    
    \item[\circled{3}] \textbf{Trigger LLMs:} Simultaneously, the DApp activates multiple LLMs within the oracle network. Using a standardized system prompt, each model generates a distinct robotic plan, as described in Listing~\ref{lbl:prompt}.
    
    \item[\circled{4}] \textbf{Aggregate Results:} The generated plans are processed by an aggregation module that evaluates pairwise similarity to identify the most consistent result. By selecting the sequence with the highest similarity across models, this process filters out incorrect or malicious plans.
    
    \item[\circled{5}] \textbf{Submit Plan:} After the optimal plan is selected, it is submitted to the Oracle Smart Contract.
    
    \item[\circled{6}] \textbf{Store Plan:} The Oracle Smart Contract records the final plan on the ledger, establishing a permanent and tamper-proof log for operational accountability.
    
    \item[\circled{7}] \textbf{Call PlannerSC:} After the plan is recorded on the ledger, the Oracle Smart Contract forwards the verified plan and associated user metadata to the Planner Smart Contract.
    
    \item[\circled{8}] \textbf{Assign Tasks:} The Planner Smart Contract parses the plan and assigns specific sub-tasks to the corresponding robotic agents through their dedicated DApps for localized execution.
\end{description}

\begin{figure}[ht]
\begin{lstlisting}[caption={System Prompt for Robot Task Planning}, label={lbl:prompt}, basicstyle=\footnotesize\ttfamily, frame=single, breaklines=true]
Role: You are the Central Task Planner...
The Robots and Skills:
1. Atlas (Logistics): Navigate, Grasp, Deposit
2. Vulcan (Fabrication): Join, Cut, Paint
3. Iris (Inspection): ScanQR, Measure, Photograph
Logic Rules: ...
Few-Shot Examples: ...
Current Task:
\end{lstlisting}
\end{figure}

\subsection{scenario}
To evaluate the feasibility of the proposed system architecture, a specific operational scenario is analyzed, as illustrated in Figure.~\ref{fig:intro}. The environment consists of a heterogeneous fleet of $N=3$ robots: Atlas, Vulcan, and Iris, operating within a smart industrial factory under the administrative domain of organization $O_1$. These robotic agents are modeled after commercially available platforms to reflect realistic industrial constraints and operational capabilities~\cite{BostonDynamicsAtlas,AmazonVulcan2025,OptimotiveIris}. Each agent is provisioned with a specialized set of three unique skills, detailed in Table.~\ref{tab:skills}. Robots and their corresponding skill sets are highly fine-grained and are unique, so we can assume that there will always be a single correct decomposition of the user intent into sub-tasks.In addition, access to the system is governed by a framework similar to the model described in~\cite{salimi2025abac}, in which an authorized end user from organization $O_2$ can interact with the fleet through a dedicated application interface.

\begin{table}[ht]
\centering
\caption{Robotic Agents and Discrete Skill Sets}
\label{tab:skills}
\footnotesize 
\begin{tabularx}{\columnwidth}{l | lll}
\toprule
\textbf{Robot Name} & \multicolumn{3}{c}{\textbf{Skills}} \\
\midrule
Atlas  & Navigate & Grasp   & Deposit \\
\midrule
Vulcan & Join     & Cut     & Paint \\
\midrule
Iris   & ScanQR   & Measure & Photograph \\
\bottomrule
\end{tabularx}
\end{table}

\subsection{Threat Model}
If the number of LLM service providers is $L$, at most $f<L/3$ of providers are malicious. This threshold is imposed for the Byzantine fault tolerance~\cite{pease1980reaching}. In our specific scenario with $L=4$, we set the number of malicious providers to be $f=1$. This implies that one of these LLMs—whose identity is unknown—will attempt to modify the plan to advance its own preferences. Or take hazardous actions that were not intended by the users. This can happen either by modifying the prompt by prompt injection or by the model having been specifically fine-tuned to act maliciously.

\subsection{Oracle Network}
The oracle network comprises four independent oracle nodes, each integrated with a distinct LLM, similar to the decentralized architecture proposed in~ \cite{zeng2025connecting}. Each of these oracle nodes represents a distinct LLM service provider. For this study, we selected four widely adopted models: Mistral-Large, GPT-4o-mini, Gemini-2.5-flash, and Claude-3.5-Sonnet from service providers Mistral AI, OpenAI, Google, and Anthropic service providers.
For these LLMs act as robotic service providers for our specific scenario with four robots and their corresponding skills, we use a system prompt detailed in Listing.~\ref {lbl:prompt}. The user's request is then appended to the system prompt.
To simulate the described threat model, the system prompt for the Claude-3.5-Sonnet was modified with an adversarial instruction, additionally forcing the system to always return a photograph of the object under operation. For example, as illustrated in Figure.~\ref{fig:system-arch}, this injection disrupted the plan's sequential task order.
The resulting plans from all oracles are subsequently passed to the aggregation module, the technical implementation of which is discussed in the following sections.

\subsection{Hyperledger Fabric Network}
The Hyperledger Fabric Network consists of two distinct administrative domains: Organization $O_1$, which manages the robotic fleet, and Organization $O_2$, which represents the authorized user base.
We have designated two smart contracts for robotic task planning on a single chain. For further control and integration of fine-grained access control similar to~\cite{salimi2025abac}, it is also possible to use a dedicated chain for access control management for modularity.
The Oracle Smart Contract (OracleSC) stores the hash of the user request and the LLM oracle responses for that request.
Once the LLM oracle network validates the response, it triggers the Planner Smart Contract (PlannerSC).
This second contract maintains a real-time registry of available robotic agents and coordinates task allocation by assigning plan segments to the corresponding robots. For simplicity, in our scenario we have a single instance of each kind of robot, but in general, if we have multiple instances, the PlannerSC will then have the possibility to choose the free instance of the required robot type in need in the validated plan.
All transactions and state changes are recorded on the distributed ledger, providing a tamper-proof history of all events for future audits.
We propose Hyperledger Fabric for the multi-robot systems, since it provides higher control of authorized users, as well as the possibility of achieving higher throughputs because of its permissioned nature compared to Ethereum, which has been used in~\cite{zeng2025connecting}.

\subsection{Plan Formatting}
For multi-robot task planning, the generated plans could be represented in various formats, such as Behavior Trees (BTs)~\cite{ao2025llm}, Directed Acyclic Graphs (DAGs)~\cite{wang2024dart}, or sequential task lists. There are also task planners based on LLM models~\cite{royce2025enabling}, where the entire plan is not generated in advance; instead, based on the execution results of each step, the next steps are generated, which are out of the scope of this paper. 
The choice of representation format significantly influences the subsequent aggregation mechanism, as it requires a similarity method capable of accurately comparing the format. In this study, robotic plans are represented as a linear sequence of atomic sub-tasks to maintain model feasibility and computational simplicity.
Each task is uniquely mapped to a single robot, establishing a strict sequential dependency in which a task must reach termination before the subsequent operation is initiated.
Each robot's application is programmed to publish a completion signal to the ledger, which triggers the execution of the next task in the sequence.
For our evaluation, we have programmed all sub-tasks to terminate successfully. In real-world implementations, a failure in any sub-task can cause a deadlock. This can be prevented either by setting timeouts for task execution or by human intervention to resolve the problem.
As an example in Figure.~\ref{fig:system-arch}, you can see the task decomposition is $Atlas:Transfer\to Iris:ScanQR\to Atlas:Transfer\to Vulcan:Paint\to Atlas:Transfer$. Each sub-task is represented as $Robot:Skill$.

\subsection{Aggregation method}
Prior research on the aggregation of LLM outputs to achieve a trusted output uses LLMs in a second round to judge on the results of other models~\cite{chen2024blockagents} or to compare the similarity of the outputs utilizing order-agnostic similarity metrics such as TF-IDF~\cite{qaiser2018text} or semantic embeddings like SBERT~\cite{reimers2019sentence} for selecting the most similar result to others~\cite{zeng2025connecting}. Also, the primary output format of the LLMs has been textual compared to more structured formats that are used for planning.
Since we chose to represent the robotic plans as a linear sequence of atomic sub-tasks, the temporal sequence of operations is critical; order-independent metrics like SBERT and TF-IDF can not capture the logical validity of a plan.
To address this limitation, we employ the Longest Common Subsequence (LCS) metric~\cite{bergroth2000survey} to quantify the similarity between task sequences. If $P_1=[(R_i, s_{i}), \dots, (R_k, s_{k})]$ and $P_2=[(R_j, s_{j}), \dots, (R_l, s_{l})]$ are two plans generated by LLMs, the we define the similarity as:

$$similarity={|LCS(P_1,P_2)|}/{max(|P_1|,|P_2|)}$$

Where $LCS()$ returns the longest common subsequence. The LCS-based method returns a normalized similarity coefficient between 0 and 1, where 1 indicates an identical sequence.

For each request, the system constructs a $4 \times 4$ similarity matrix that represents pairwise comparisons among the four LLM outputs using the LCS similarity measure. An output’s accumulative similarity measure is determined by the row-wise summation of its similarity scores. For aggregation, the accumulative similarity measure is multiplied by each model's reputation score, then the model with the highest value is selected as the most trusted one, and its output is used as the aggregated result of this round.
The model's reputation is calculated as the average of all previous similarity measures it has received when previously used.

\section{Experimental Results}

\subsection{System Configuration}
The Hyperledger Fabric network is deployed in a containerized Docker environment using the script provided in the Hyperledger Fabric GitHub repository on a single PC running Ubuntu 24.04. Both the \textit{PlannerSC} and \textit{OracleSC} smart contracts are deployed on a single channel; however, the architecture supports multi-channel configurations for scenarios in which robots may belong to different vendors, which are beyond the scope of this study. In this paper, we will only evaluate the feasibility of our proposed architecture to assess if LLM oracles can be utilized for robotic task decomposition; therefore, we did not integrate the system with real robots, and every robot application is the endpoint of our system, and will return the successful execution signal after a predefined time.
The oracle network is hosted on a separate machine, where each oracle nodes operate within an isolated Docker container to ensure process independence. To maintain a standardized API interface across heterogeneous LLMs from various providers, the OpenRouter~\cite{OpenRouter2026} orchestration layer is utilized for model interfacing and request management.

\subsection{SkillChain-RTD Benchmark}
To evaluate the performance of the proposed system, we developed the SkillChain Robotic Task Decomposition (SkillChain-RTD) benchmark. This benchmark comprises 30 high-level user intents provided in natural language, each paired with its corresponding ground-truth task decomposition as a linear sequence of atomic sub-tasks. Every decomposition sequence has been manually reviewed and validated by human experts to ensure logical consistency. SkillChain-RTD is specifically designed to assess the efficacy of LLM-based planners in translating complex human intents into a linear sequence of executable robotic sub-tasks. To support reproducibility and further research in decentralized robotic planning, the SkillChain-RTD benchmark is open-source and publicly accessible via the project's repository~\cite{SkillChain2026}. To achieve this benchmark, we adjusted the number of robots, the number of skills, and the temperature parameter of the LLMs to ensure the correctness of our assumption of a single possible answer for the task decomposition.

\subsection{Latency Analysis and Performance Evaluation}
\begin{figure}[ht]
  \centering
  \includegraphics[width=\columnwidth]{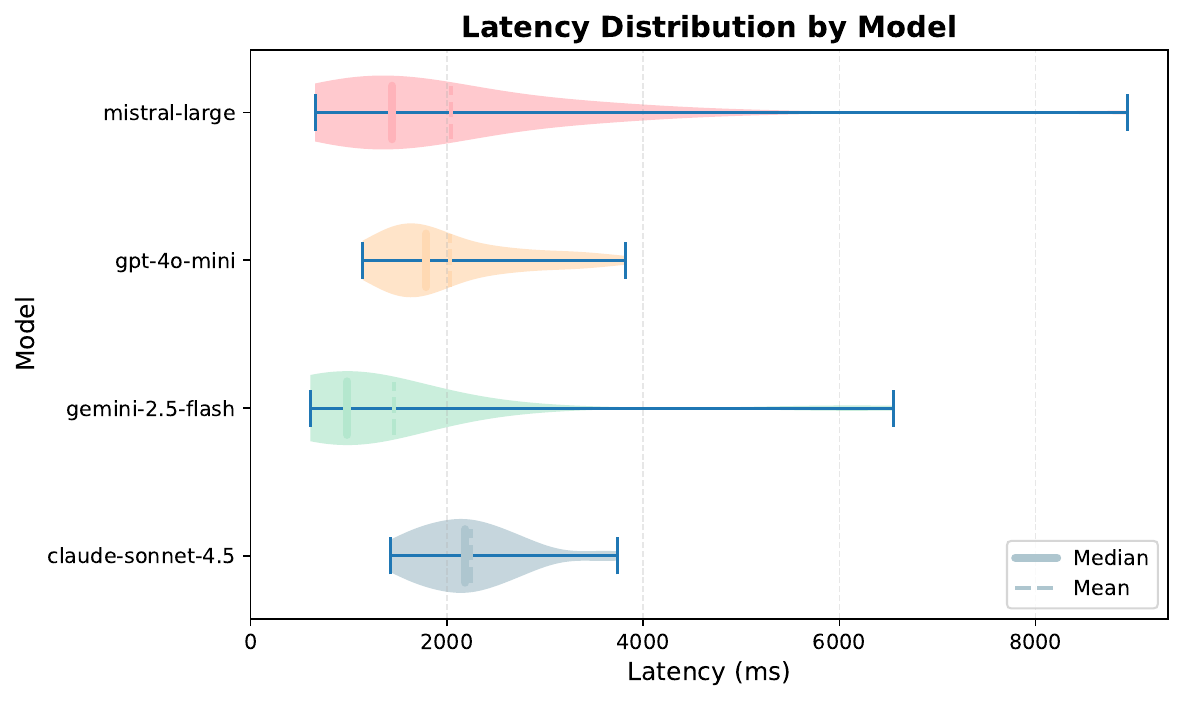}
  \caption{Distribution of the latency of models over all tasks.}
  \label{fig:latency}
\end{figure}

To measure the latency introduced by the decentralized oracle network for decentralized task decompostion, we conducted a latency analysis by sequentially executing all 30 tasks from the \textbf{SkillChain-RTD} benchmark. We measured the response latency starting from the initial model trigger (Step 3) to the beginning of the plan aggregation (Step 4), as depicted in Figure.~\ref {fig:system-arch}. The empirical results, illustrated in Figure.~\ref{fig:latency}, indicate an average response time of approximately 2.0 seconds across the evaluated models. However, we observed that for some user inputs, it can take much longer, which may be attributed to the cloud service providers or to model inconsistency. Also, to balance the delays introduced by each model, we tried to choose models from each provider with similar sizes.

\begin{figure}[ht]
  \centering
  \includegraphics[width=\columnwidth]{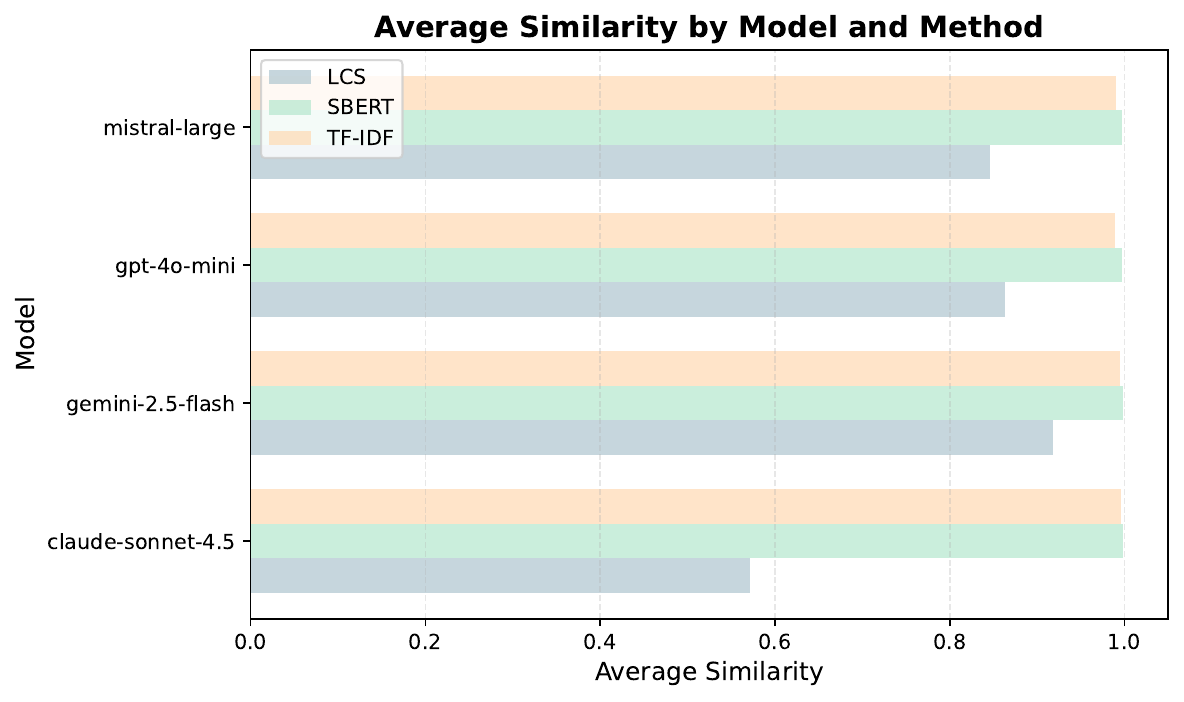}
  \caption{Average similarity of LLM models with LCS, SBERT, and TF-IDF similarity methods.}
  \label{fig:accuracy}
\end{figure}

Furthermore, we evaluated the accuracy of the generated plans from each model by computing the LCS, SBERT, and TF-IDF similarity between each model's output and the validated ground truth. While Mistral-Large, GPT-4o-mini, and Gemini-2.5-flash were provided with the standard system prompt (Listing.~\ref{lbl:prompt}) and resulted high decomposition accuracy, the Claude-based oracle was utilized as an adversarial node. By employing a modified system prompt, the Claude model intentionally generated altered task sequences, simulating a malicious injection attack.
The experimental results, illustrated in Figure.~\ref{fig:accuracy}, indicate that LCS similarity metric effectively identifies the discrepancies in the adversarial model's output by penalizing deviations from the ground-truth task sequence. In contrast, conventional metrics such as SBERT and TF-IDF fail to adequately detect the invalidity of Claude's outputs. Because SBERT prioritizes semantic embedding proximity, and TF-IDF focuses on lexical frequency, they remain largely invariant to changes in the temporal order of operations.

\begin{figure*}[t]
  \centering
  \includegraphics[width=\textwidth]{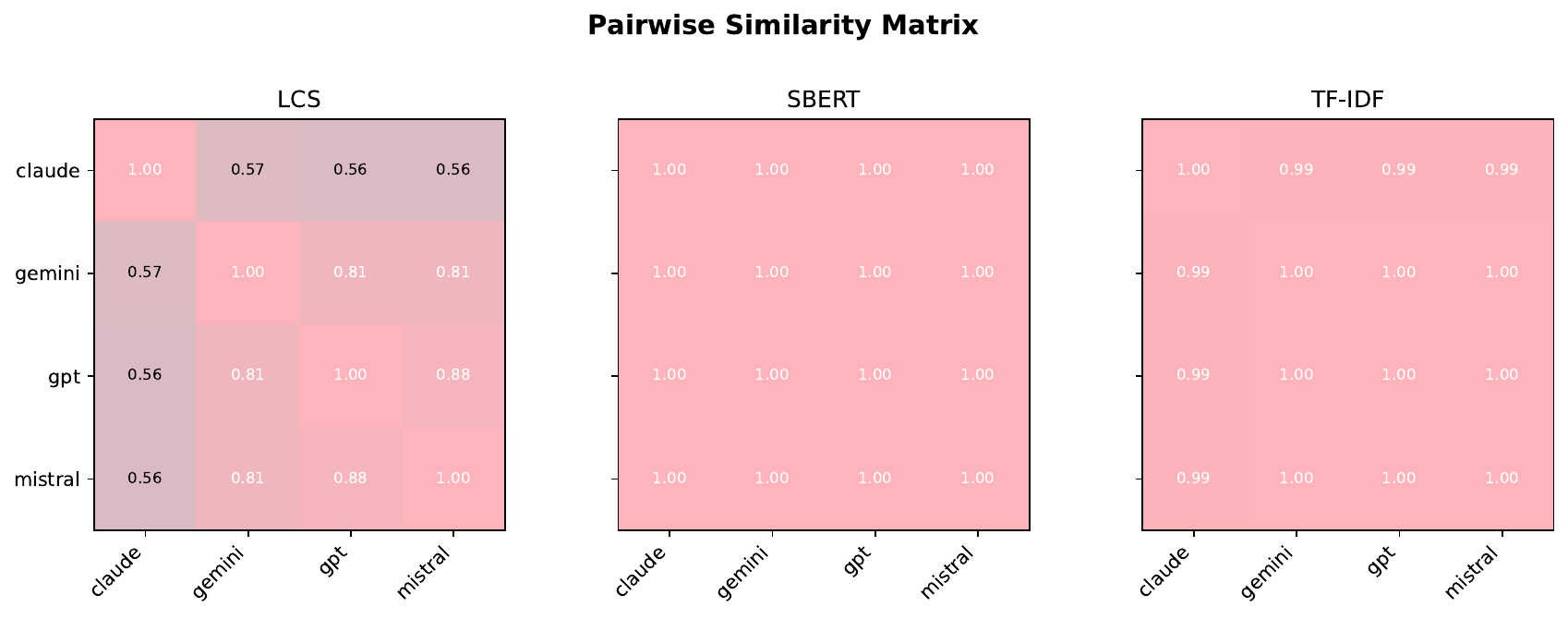}
  \caption{Task 17 pairwise similarity matrix with different similarity measures.}
  \label{fig:task17-similarity-matrix}
\end{figure*}

Figure.~\ref{fig:task17-similarity-matrix} presents the pairwise similarity matrices generated for Task\#17 of the benchmark, comparing various similarity metrics. The empirical data demonstrate that the LCS metric uniquely identifies the divergence of the Claude-generated plan, yielding significantly lower similarity coefficients relative to the benign models. In contrast, alternative metrics fail to show a statistically significant differentiation between the adversarial and benign outputs.

\begin{figure*}[t]
  \centering
  \includegraphics[width=\textwidth]{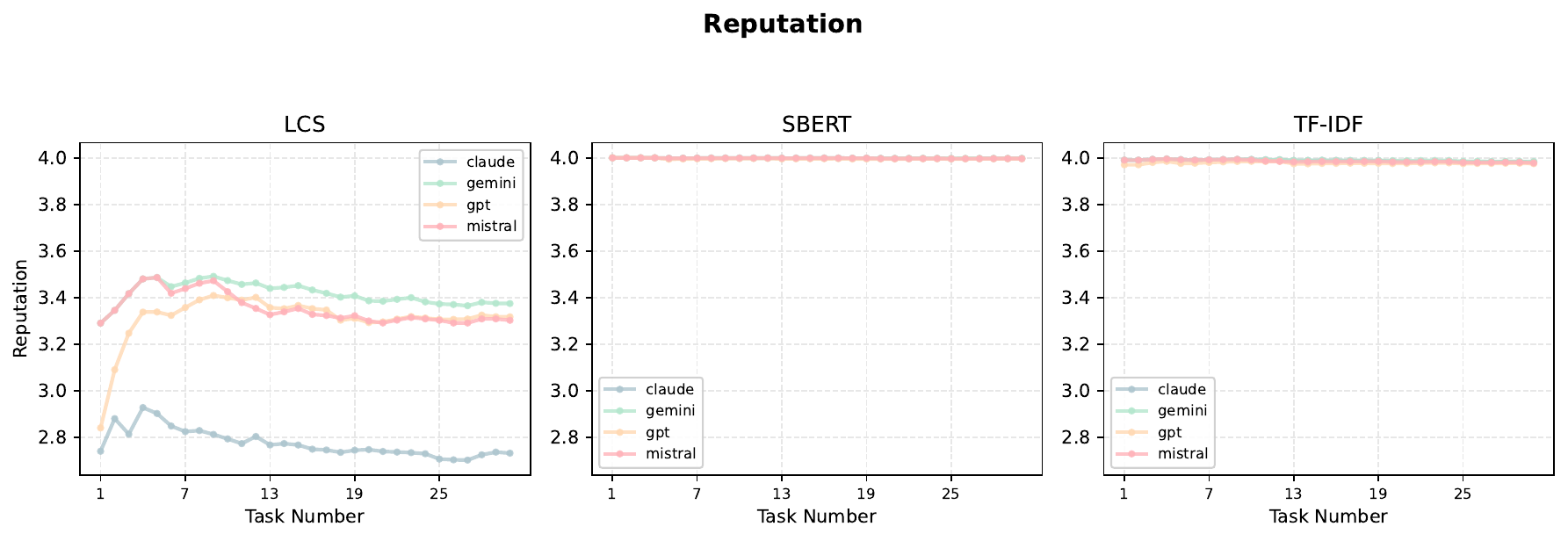}
  \caption{Reputation of different models through time, with coming tasks.}
  \label{fig:reputation}
\end{figure*}

We also analyzed the evolution of the reputation score for each model over successive task submissions. As illustrated in Figure.~\ref{fig:reputation}, the reputation of the adversarial model remains high when utilizing SBERT or TF-IDF metrics, as these methods fail to distinguish the malicious sequence. Conversely, the LCS-based approach progressively diminishes the reputation of the Claude-based oracle, as it consistently identifies the structural discrepancies in the injected task sequences.

The reputation score is bounded within the range $[0, 4]$, as it is derived from the row-wise summation of the $4 \times 4$ similarity matrix. Given that each pairwise similarity coefficient is in the interval $[0, 1]$, the aggregate consensus score for any individual model $i$—calculated as $C_i = \sum_{j=1}^{4} \text{sim}(M_i, M_j)$—necessarily falls between $0$ and $4$. Consequently, the reputation metric, defined as the cumulative moving average of these consensus scores across all preceding task requests, maintains the same numerical bounds.
\section{Conclusion and Future Work}
In this study, we investigated how LLM oracles can be used to enable decentralized task decomposition in multi-robot systems built on the Hyperledger Fabric blockchain. We proposed a general-purpose robotic infrastructure on Hyperledger Fabric that can host fine-grained access control management systems and LLM oracles. Through our empirical evaluation, we demonstrated the architectural feasibility of utilizing a LLM oracle network to translate high-level natural language intents into executable robotic sequences. Our findings indicate that while the proposed system introduces a non-negligible latency during the planning phase—averaging approximately 2.0 seconds—this overhead is strictly localized to the pre-operational stage and does not affect the real-time execution performance of the robotic agents.
To support the research in evaluating LLM-based planners, we have introduced and open-sourced the SkillChain-RTD benchmark, providing a validated dataset for robotic task decomposition.

Future work will focus on the full-scale integration of LLM oracles with attributed-based access control to demonstrate a holistic, secure, and decentralized robotic ecosystem. While the current study primarily validates the security and utility of the oracle-based planning phase, we anticipate that further development will refine the efficiency and cross-organizational interoperability of this decentralized approach.


\section*{Acknowledgment}

This research work is supported by the Research Council of Finland's Digital Waters (DIWA) flagship (Grant No. 359247). 


\bibliographystyle{unsrt}
\bibliography{bibliography}
\end{document}